# Neural Mechanism of Language

Peilei Liu[1]*, Ting Wang[1]

**Affiliations:**

[1]College of Computer, National University of Defense Technology, 410073 Changsha, Hunan, China.

*Correspondence to: plliu@nudt.edu.cn

**Abstract**: This paper is based on our previous work on neural coding. It is a self-organized model supported by existing evidences. Firstly, we briefly introduce this model in this paper, and then we explain the neural mechanism of language and reasoning with it. Moreover, we find that the position of an area determines its importance. Specifically, language relevant areas are in the capital position of the cortical kingdom. Therefore they are closely related with autonomous consciousness and working memories. In essence, language is a miniature of the real world. Briefly, this paper would like to bridge the gap between molecule mechanism of neurons and advanced functions such as language and reasoning.

**Main Text:** Language is an advanced intelligence almost peculiar to human. We wish to explain the neural mechanism correlative with language in this paper. Namely to answer the following questions: How do we make sentences those we have never heard of? And why language is important? Language here means speaking language, while written language is very similar.

Detailed models of synapse, neuron, and lateral competition were introduced in our previous work (*1-3*). Briefly, neural circuits are self-adaptive to external inputs following several rules (R): R1) neurons firing meanwhile tend to be connected by the same neuron (*1*). R2) Neurons firing earlier tend to connect to neurons firing latter, while the latter tend to inhibit the former through LTD (long term depression) or inhibitory intermediate neurons (*1, 3*). Both the strengthening and inhibition are negative correlative with the time interval between neurons' firing. This is similar to the STDP (*4*), but STDP ignores the temporal summarization of spikes (*1*). Incidentally, all these rules are merely tendencies following statistical laws. R3) The neuron is an function of all inputs: $f = c_1(1-e^{-c_2\sigma})$ where $\sigma = \sum_i f_i$, f and $f_i$ are firing frequencies (*1*). R4) There is lateral inhibition between neurons sharing common inputs (*1*). Every picture is encoded by a single neuron in this model, and the neuron's firing frequency represents the probability estimate of this picture occurring (*1*). Specially, a motion or action is also encoded by a single neuron (*2*).

A sentence in essence is a temporal sequence of words. And every word should correspond to a single neuron according to our model, whether it is noun, verb or others. A temporal sequence can be encoded by self-organization of neurons according to R2 (see Fig. 1A). Specifically, previous coding neurons tend to connect to succeeding ones, while succeeding neurons tend to depress previous ones. As in our previous work on vision (*2*), motion in the retinas is a sequence on the milliseconds scale in essence. And it is actually encoded by synapses with time delay. Similarly, the natural delay between a neuron's output and input can be used for encoding sequences in the cortex (see Fig. 1A). Moreover, a circuit and hormones accumulation has longer delay than a single neuron. In essence, these delays are natural timers, just like a clock's different hands. And sequence coding on different scales is actually implemented through these timers. For example, the biological clock in essence is a sequence coding on the hours scale.

On the other hand, sequence coding is a kind of association. Connections between neurons in a sequence can be viewed as ordinary inputs. In theory, even a single dendritic input can fire the whole neuron according to R3. Therefore an external input would stir up many neurons. However, the actual fired neurons at any time are actually determined by lateral inhibition according to R4 (*1*). When the number of dendritic inputs is very small, the neuron's firing is actually association. Otherwise when the number of dendritic inputs is very large, the neuron's firing is usually called recognition or identification. In other words, association and recognition are similar. And the difference lies in the number of retrieval clues. A 3D (three-dimensional) object can be encoded by neurons with such associations. For example, a grandmother has many pictures from various angles, each of which is encoded by a single neuron. These neurons associate with each other and compose a sequence or circuit, which actually encoding this grandmother. Therefore a 3D object can be represented by a series of associated 2D pictures. The number of these pictures determines the distinguishability of this object. The initial or most common angle such as the front view will have the strongest connections with others. And therefore it can somewhat represent the grandmother as a classic or standard impression. This has reconciled the "grandmother cell" and population coding in some degree (*5*). In our opinion, feature constancy should result from similar mechanism (*2*). Moreover, if time were viewed as the fourth dimension of space, a temporal sequence is similar to a 3D object. In fact, various viewpoints of a 3D object indeed enter the cortex one by one successively. Therefore associations between them can form according to R2 (see Fig. 1A).

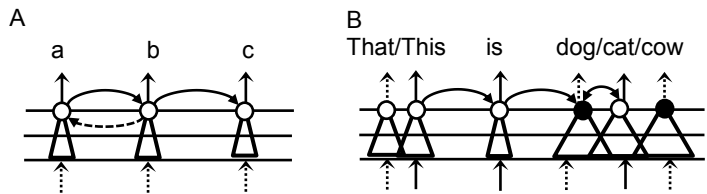

**Fig. 1. Sequence coding and language.** Triangles represent coding trees (*1*). White and black dots represent fired and unfired neurons respectively. Solid vertical arrows are actual inputs and outputs, while dotted-line arrows are possible ones.  In A, sequence coding is self-organization following R2. Specifically, connection from a to b is strengthened, while connection from b to a is weakened through LTD. In B, language is the result of sequence coding and lateral inhibition. Accusative "dog", "cat" and "cow" compete with each other, because they share some common inputs or features.

A little girl can make various sentences that she has never heard of. In fact, this can be implemented as following. Suppose she has learned the sentence "this is dog" (see Fig. 1B). Although neuron "is" connects to "dog", it doesn't mean that "is" will fire "dog" absolutely. Whether neuron "dog" would be fired is actually determined by lateral inhibition. Specifically, neurons "dog", "cat" and "cow" will compete with each other, because they have common dendritic inputs which representing features. The neuron with the most inputs will win and be selected. Therefore, if the girl is just seeing a cat picture or thinking of a cat from other clues at that time, she will select the word "cat" instead of dog. And then she will make a new sentence "this is cat". Similarly, other sentence elements such as subject and predicate can also be selected in this way. Since most nouns could be both subjects and objects, there should be abundant circuits. On the other hand, many researchers worry that neurons in the cortex are too few for encoding the infinite number of sentences. In general, every word is encoded by a sing neuron, because it is approximately transient similar to a picture. A temporal sentence however corresponds to a timeline. Therefore it should be encoded by a sequence of neurons other than single neuron. In theory therefore, tens of thousands neurons (equal to the number of words) are enough for encoding infinite sentences. In fact, the initial and frequently-used sentences are actually viewed as sentence patterns by the cortex. Specific words in every position are mutually-exclusive due to lateral inhibition. And every word is selected according to external and internal inputs at that time. In whole, the cortex is a "Turing Machine" in essence, whose current output is determined by current external inputs and internal state.

Another talent of human is reasoning or logical inference. As is well known, the combination of logic operations "IMP" ('a implies b' or "a->b") and "NOT" is computationally complete (*6*). Operation "IMP" is actually a short temporal sequence which can be implemented as mentioned above (see Fig. 1). Operation "FAUSE" and "NOT" can be implemented by lateral inhibition or inhibitory intermediate neurons (*1, 3*). Therefore our model can execute any logical inference and reasoning (*1, 3*). Moreover, this inference is transitive, namely rule 'a->b' and 'b->c' together can result in rule 'a->c'. Specifically, with synapses growing, the time delay between a and c will become smaller. Since connection strength is negative correlative with time interval according to R2, a tends to connect to c directly. Briefly, reasoning process in the cortex is companied with the contraction of pathways and the formation of new associations. And as the old said, practice makes perfect. From this viewpoint, the cortex is actually a repository containing a huge number of such "IMP" equations. And the activation of an existing circuit corresponds to an interference process. Most of all, new rules will be generated automatically in this process, and contradictory rules will be reconciled gradually.

Since the neural network is self-similar according to our model (*1*), any small area can be viewed as a mini cerebrum. And fibers casting to this area can be viewed as external inputs. According to R3, a neuron is the exponential summarization of all inputs other than linear summarization. Therefore inputs actually converge and shrink (see Fig. 2) (*3*). And direct inputs have greater influence than indirect inputs. Specifically, the influence of an input is negative correlative with its logic distance (the number of relaying neurons). As results, neurons far away from external inputs are rarely influenced by them. Therefore they are relatively autonomous and free. On the other hand, the output end is divergent (see Fig. 2) (*3*). Therefore the neuron farthest away from the ultimate output is most powerful, because it can influence most motor units. As shown in Fig. 2, there should be some autonomous as well as powerful neurons existing in the kernel area (*2*). From the physiological evidences, this area should be PFC (prefrontal cortex) or the frontal lobe. It is like the capital of the cortex kingdom, where language and declarative memory is stored. If an input isn't strong enough to enter this capital, it can only influence subconsciousness and implicit memory (see Fig. 2). In fact, many factors influence this process such as: the strength of input signals, the length of signal duration, lateral competition with the internal signals, and neural hormones and modulators. The neural network is somewhat like roads networks: most country roads lead to the city, but roads in the city compose complex circuits. Moreover there are some long fibers across different areas, just like expressways between cities. This main structure is determined by genes according to Crick (*7*), while specific fine connections are determined by postnatal experiences (*1*).

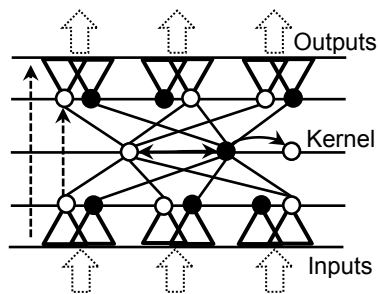

**Fig. 2. Declarative memory and consciousness.** You can get this figure through cutting and overlapping cortical areas one above another successively. White and black dots mean fired and unfired neurons respectively. The whole neural network is actually composed of many overlapping sandglass structures. And neurons in their waists compose the kernel, which approximately corresponds to the PFC area. The kernel is autonomous and powerful due to their central positions, and there are abundant circuits here. Signals within these circuits represent our autonomous consciousness. Declarative memory and working memory are synaptic long-term and short-term strengthening respectively here. Dotted-lines represent implicit memory. Correspondingly, signals within them represent subconsciousness or unconsciousness. It is hard for us to perceive them, because they are far away from the capital and language areas.

As widely acknowledged in neuroscience, memories are stored in synaptic connections. According to Crick, consciousness should be a kind of instantaneous memories (*7*). According to our model (*1*), consciousness is signals within circuits while working memory is short-term strengthening of corresponding synapses. In theory, every firing neuron has consciousness. Due to lateral inhibition however, only few neurons can be fired at any time. Therefore our consciousness is somewhat unified. In other words, our consciousness only exists in a small circuit in the cortex at any time. But it could shift between different areas with working content and attention. For example, when seeing pictures and listening music, signals should exist in different circuits and areas. Attention is actually some neural hormones or modulators such as norepinephrine (*1*). It is essential for maintaining signal circulation, just like gasoline for internal-combustion engine. The pervasive casting of these neuromodulators will influence the result of lateral inhibition between areas. For most people however, consciousness exits in language relevant areas at most times. Since oral reports were usually used as the results of consciousness relevant test, this impression was strengthened. Some people even insist that consciousness as well as language is peculiar to human. This is equivocal for the deaf-mutes, babies and some normal people such as painters. When working, painters could use pictures instead of words for thinking. According to our model, almost all animals have consciousness, just like every country having a capital whether it is large or small. Even if this capital is destroyed in accident, another great city will be selected as the new capital. Moreover, when the capital has a rest in sleep, dreams will emerge spontaneously due to the release of depression from consciousness (*1*). That might be why dream can reflect subconsciousness and unconsciousness (see Fig. 2). However, consciousness of animals is indeed not as autonomous and free as human, because the capital is close to sense organs due to their cortex size.

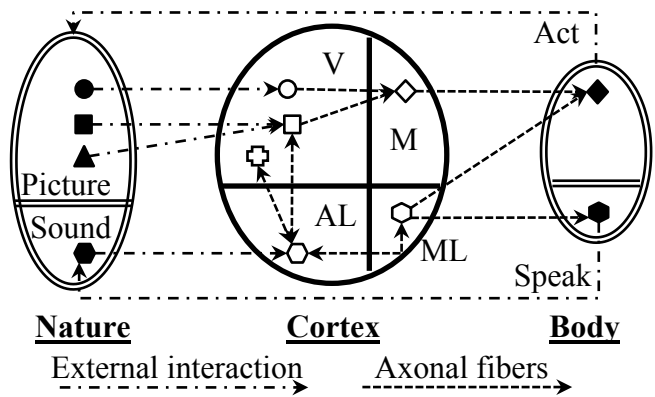

**Fig. 3. World model of language.** Black shapes represent objects or muscles, while white shapes mean neurons. The cortex is divided into four parts: V (visual area), AL (auditory language area), M (motor area), and ML (motor language area). V is a "mult-to-one" map of the nature, while AL is a "mult-to-one" map of V. Therefore V actually composes the world model of AL, and its changes will result in the changes of AL. Reversely, AL can react to V and the nature. AL is the capital of the cortex for most people. The hexagons and relevant pathways here approximately correspond to declarative memory and consciousness. Shapes and pathways in V and M approximately correspond to implicit memory and subconsciousness.

When learning language, we actually build a "multi-to-one" correspondence from objects and motions to words (see Fig. 3). Briefly, a word's connections with pictures determine its semantic meaning, while connections with other words determine its syntax role. Therefore, these coding neurons in visual area actually compose the "world model" of language. From this viewpoint, the language relevant area is the brain in brain. If the cortex is called the second nature (*1*), language relevant area is the third nature. Specifically, neurons correspond to objects, attributes and motions, while connections correspond to associations and temporal relations. Therefore language could be the key to consciousness research. The syntax structure of a language actually reflects the actual cortical circuits of the speaker. Generally speaking, that we can make various new sentences is because the continually changes of the real world result in the changes of the language space (see Fig. 3). And reversely, language can react to one's mind and the real world as well. In hypnosis for example, language is used to change the consciousness. And patients are usually required to close eyes and keep relax, because external inputs will compete for influence on consciousness with the language of hypnotist. And this can also explain why closing eyes helps in deeply thinking. The cortex and language are actually a simulation of the world. And imagination and plans are a kind of preview in this "virtual" space. From the evolution viewpoint, the generation of language is due to the need of communication. Ultimately, only external inputs can change one's mind or consciousness. And we can make these external inputs artificially through actions. Compared with drawing pictures or making gestures however, making pronunciations has advantages in easiness and propagation distance. That might be why we use language as the tool of thinking and communication.

**Acknowledgments:** This work was supported by the National Natural Science Foundation of China (Grant No. 61170156 and 60933005).